\documentclass[runningheads]{llncs}

\usepackage{eccv}

\usepackage{eccvabbrv}

\usepackage{graphicx}
\usepackage{booktabs}
\usepackage[misc,geometry]{ifsym}
\usepackage{caption}
\usepackage{subcaption}
\usepackage{multirow} 

\usepackage[accsupp]{axessibility}  

\usepackage{hyperref}

\usepackage{orcidlink}

\begin{document}

\title{WAS: Dataset and Methods for Artistic Text Segmentation} 

\titlerunning{WAS: Dataset and Methods for Artistic Text Segmentation}

\author{Xudong Xie\inst{1} \and
Yuzhe Li\inst{1} \and
Yang Liu\inst{1} \and
Zhifei Zhang\inst{2} \and
Zhaowen Wang\inst{2} \and
Wei Xiong\inst{2} \and
Xiang Bai\inst{1}\textsuperscript{(\Letter)}}

\authorrunning{X.~Xie et al.}

\institute{Huazhong University of Science and Technology, China
\email{\{xdxie,yzli12,yangliu1213,xbai\}@hust.edu.cn} \and
Adobe, USA \\
\email{\{zzhang,zhawang\}@adobe.com}, \email{wxiongur@gmail.com}}

\maketitle

\def\thefootnote{\Letter}\footnotetext{Corresponding author}\def\thefootnote{\arabic{footnote}}

\begin{abstract}
  Accurate text segmentation results are crucial for text-related generative tasks, such as text image generation, text editing, text removal, and text style transfer. Recently, some scene text segmentation methods have made significant progress in segmenting regular text. However, these methods perform poorly in scenarios containing artistic text. Therefore, this paper focuses on the more challenging task of artistic text segmentation and constructs a real artistic text segmentation dataset. One challenge of the task is that the local stroke shapes of artistic text are changeable with diversity and complexity. We propose a decoder with the layer-wise momentum query to prevent the model from ignoring stroke regions of special shapes. Another challenge is the complexity of the global topological structure. We further design a skeleton-assisted head to guide the model to focus on the global structure. Additionally, to enhance the generalization performance of the text segmentation model, we propose a strategy for training data synthesis, based on the large multi-modal model and the diffusion model. Experimental results show that our proposed method and synthetic dataset can significantly enhance the performance of artistic text segmentation and achieve state-of-the-art results on other public datasets. The datasets and codes are available at: \href{https://github.com/xdxie/WAS_WordArt-Segmentation}{https://github.com/xdxie/WAS\_WordArt-Segmentation}.
  \keywords{ artistic text segmentation \and momentum query \and skeleton}
\end{abstract}

\section{Introduction}
\label{sec:intro}
Text segmentation is dedicated to finely segmenting the strokes of text from complex scene images, discriminating whether each pixel belongs to the text foreground or the background. Accurate text segmentation results are the foundation for text-related generative tasks. For instance, tasks such as text image generation\cite{NEURIPS2023_1df4afb0}, text style transfer~\cite{li2020fet,lyu2017auto}, and text removal~\cite{wang2023real,10.1145/3343031.3350929} can produce excellent and practical generative outcomes based on text masks. However, although existing models have achieved outstanding performance in regular text segmentation tasks, they are difficult to accurately segment artistic text. Artistic text features complex appearances and shapes, making it difficult to distinguish from background patterns~\cite{xie2022toward}. Therefore, artistic text segmentation is a challenging task that has not yet been studied by the academic community.

The first problem encountered in implementing this new task is the lack of datasets. The existing text segmentation datasets ICDAR13 FST~\cite{icdar13}, COCO\_TS\cite{cocots}, MLT\_S~\cite{mlts}, and Total-Text~\cite{totaltext} suffer from the problems of low annotation quality and insufficient quantity. Moreover, these data are scene images with regular text. Although TextSeg~\cite{xu2021rethinking} provides high-quality annotated data and some artistic text images, the number of these images is still insufficient to train a high-performance artistic text segmentation model with strong generalizability. Therefore, we construct a real \textbf{W}ord\textbf{A}rt \textbf{S}egmentation dataset called \textbf{WAS-R}, consisting of 7100 artistic text images with word-level annotations of quadrilateral boxes, masks, and transcriptions. Additionally, to further enhance the accuracy and generalization ability of the text segmentation model, we also propose a synthetic dataset called \textbf{WAS-S}. Our designed synthetic pipeline utilizes the popular large multi-modal model and diffusion model to achieve realism, accuracy, and diversity in the generated images.

Besides, artistic text segmentation presents two unique challenges compared to general object segmentation and regular text segmentation. (1) The strokes of artistic text have flexible and changeable local shapes, such as slender tails or twisted ligatures. (2) The global topological structure of the artistic text is very complex, with many holes and intricate connections within the text. In contrast, the local stroke shapes and the global structure of regular text are almost invariant, and the topological structure of general objects is very simple. Therefore, the task we propose has clear academic value and practical significance. 

There are currently few specialized models for text segmentation. Recent studies either require the aid of text detection modules~\cite{xu2022bts,yu2023scene} or the assistance of character-level recognizers~\cite{xu2021rethinking}.
Moreover, these methods have not been specifically designed for artistic text. In view of this, we propose a WordArt segmentation model \textbf{WASNet}. To address the first challenge, we propose a Transformer decoder with the layer-wise momentum query. The input of the self-attention module is the momentum superposition of the masked queries from the current layer and the previous layers. This operation ensures that the model does not quickly fit a restricted regular mask area when updating attention, thereby ignoring some special-shaped stroke regions, which are precisely what the earlier layers are capable of capturing. To address the second challenge, we propose a skeleton-assisted head that enables the model to output both mask predictions and skeleton predictions simultaneously, guiding the model to capture the global topological structure. It enhances the decoder's ability to perceive the overall structure of the text.

We conduct extensive experiments to verify the effectiveness of the proposed method and the synthetic dataset on the task of artistic text segmentation. We also verified the generalizability on other public datasets~\cite{cocots,totaltext,xu2021rethinking}.
The results achieved state-of-the-art (SOTA) performance. More importantly, the model trained on the WAS dataset can be directly tested on other datasets without the need for fine-tuning and still achieve competitive results. This opens up a new experimental paradigm for the task of text segmentation.

In summary, our contributions are four-fold:
\begin{enumerate}
\item[(1)] We present a new challenging task: artistic text segmentation, and construct a real dataset to benchmark the performance of various models.   
\item[(2)] We design a training data synthesis strategy and generate a synthetic dataset consisting of 100$k$ image-mask pairs.
\item[(3)] We introduce the layer-wise momentum query to handle the changeable local strokes and skeleton-assisted head to capture the complex global structure.
\item[(4)] We achieve new SOTA results in the tasks of artistic text segmentation and scene text segmentation, and simplify the experimental paradigm for text segmentation.
\end{enumerate}

\section{Related Work}
\label{sec:2}

\subsection{Text Segmentation Method}

Some early text segmentation methods relied on thresholding~\cite{tra_texseg_thr2}, low-level features~\cite{tra_texseg_colr}, or Markov Random Fields (MRF)~\cite{mrf_texseg} to segment foreground text from the background, but these methods could only achieve limited success in document processing. 
With the continuous advancement of deep learning, the corresponding text segmentation methods have shown great potential in complex scenes~\cite{cnn_texseg_ccnn, cnn_texseg_nstd, mlts}. For instance, SMANet~\cite{mlts} employs the encoder-decoder architecture of PSPNet~\cite{pspnet} and achieves a multi-scale attention module for text segmentation. PGTSNet~\cite{xu2022bts} employs a pre-trained detector to ground out text regions before segmentation, further enhancing the accuracy of segmentation. TexRNet~\cite{xu2021rethinking} incorporates character recognition and attention-based similarity checking to aid the model in segmenting text. Building on these methodologies, Yu \etal~\cite{yu2023scene} developed a model featuring a lightweight detection head and a Text-Focused Module, elevating text segmentation performance in complex scenes to a new level. Nevertheless, recent high-performance text segmentation methods either utilize extra bounding box annotations and rely on text detection, or employ character-level supervision. Also, these methods lack specialized design for artistic text.

\subsection{Text Segmentation Dataset}

The construction of text segmentation datasets has not received enough attention in academia. ICDAR13~\cite{icdar13} and Total-Text~\cite{totaltext} provide high-quality, pixel-level annotations for text segmentation, but their quantities are very limited with only 462 and 1,555 images, respectively. To address the issue of insufficient quantity, researchers have proposed a dataset COCO-TS~\cite{cocots} (14,690 images) based on COCO-Text~\cite{cocotext} for text segmentation. Similarly, MLT\_S~\cite{mlts} (6,896 images) is also a large-scale text segmentation dataset based on ICDAR MLT~\cite{8270168}. Both of these datasets use automatic annotation strategies, resulting in low-quality dataset annotations. In view of these problems, Xu \etal~\cite{xu2021rethinking} introduced a larger-scale and high-quality text segmentation dataset TextSeg (4,024 images), which includes character-level and word-level annotations of masks, bounding boxes, and transcriptions. Moreover, different from the datasets mentioned above, BTS~\cite{xu2022bts} is a bilingual text segmentation dataset. These real datasets are all derived from natural images but lack a dedicated dataset for segmenting artistic text. Although TextSeg contains some artistic text images, the quantity is insufficient to train a robust model for artistic text segmentation.

\subsection{Segmentation Dataset Generation}
The construction of synthetic segmentation datasets plays an important role in studying visual perception problems. DatasetGAN~\cite{DBLP:conf/cvpr/ZhangLGYLB0F21} and BigDatasetGAN~\cite{DBLP:conf/cvpr/LiLKKF022} only use a small number of manually labeled samples for each category to train the decoder and generate a large amount of new data. 
Diffumask~\cite{DBLP:conf/iccv/WuZSZS23} extends the text-driven image synthesis to semantic mask generation in Stable Diffusion~\cite{DBLP:conf/cvpr/RombachBLEO22} to create a high-resolution and class-discriminative pixel-wise mask. 
Dataset diffusion~\cite{DBLP:conf/nips/0001VTN23} leverages the pre-trained diffusion model and text prompts to generate segmentation maps corresponding to synthetic images. 
DatasetDM~\cite{DBLP:conf/nips/WuZCGZHZSS23} is a generic dataset generation model that decode the latent code of the diffusion model as accurate perception annotations. 
MosaicFusion~\cite{DBLP:journals/corr/abs-2309-13042} is a diffusion-based data augmentation method that does not require training and does not rely on any label supervision, especially for rare and new categories. 
Then SegGen~\cite{DBLP:journals/corr/abs-2311-03355} integrates Text2Mask and Mask2Img synthesis to generate training data, improving the performance of state-of-the-art segmentation models in various segmentation tasks. 
These advanced methods often fail to align artistic text with their masks in images. In this paper, we avoid having the model generate both images and mask annotations. Instead, we pre-render the text masks and use ControlNet~\cite{DBLP:conf/iccv/ZhangRA23} to generate mask-conditioned images.

\section{Dataset}
As artistic text in the real world is incredibly diverse, we propose two new datasets: WAS-R composed of real-world text images, and WAS-S composed of synthetic text images. These multi-purpose artistic text datasets aim to bridge the gap between artistic text segmentation and real-world applications, accommodating the rapid advances in text vision research.

\subsection{WAS-R Image Collection}
The WAS-R dataset is composed of 7,100 images sourced from a variety of contexts, including posters, cards, covers, logos, goods, road signs, billboards, digital designs, and handwritten text. Among these, 4,100 images serve as the training dataset, while the remaining 3,000 images constitute the test dataset. The artistic text can be categorized into two major types according to the way capturing images. A type of artistic text image is taken by cameras from various scenes, such as signboards. The other type is directly exported from design software, such as poster files.
During data collection, we specifically balances these two types to create a diverse dataset for research and development.

\subsection{WAS-R Image Annotation}
\begin{figure}
\centering
\includegraphics[width=1\textwidth]{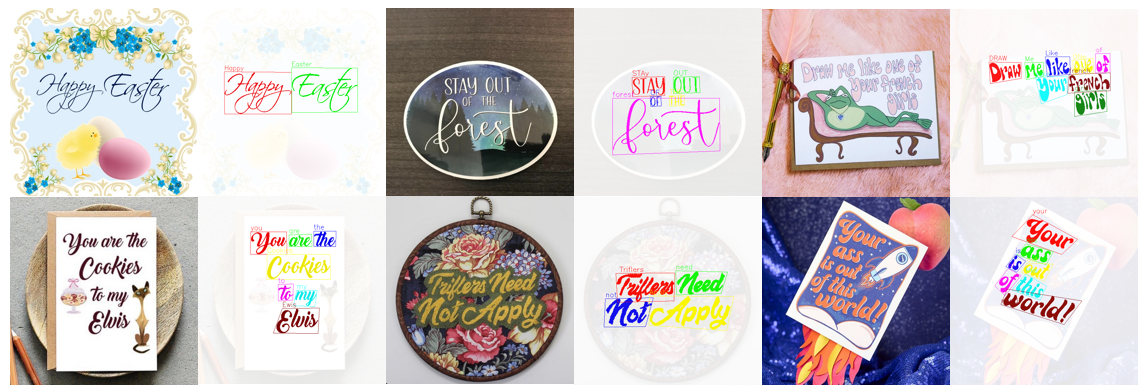}
\caption{Examples of images and annotations from the proposed WAS-R dataset. }
\label{fig:annotation}
\end{figure}

The WAS-R dataset stands out due to its comprehensive annotations, surpassing existing datasets. Specifically, WAS-R provides minimum quadrilateral detection boxes with distinct segmentation mask labels for each word. It also provides text transcription for each word mask. Moreover, we annotates the word effects such as shadow, glow, 3D, which play a crucial role in distinguishing artistic text from conventional scene text and significantly impacts text segmentation. Fig.~\ref{fig:annotation} shows examples of collected images and their annotations in WAS-R. 

\begin{figure}
\centering
\includegraphics[width=1\textwidth]{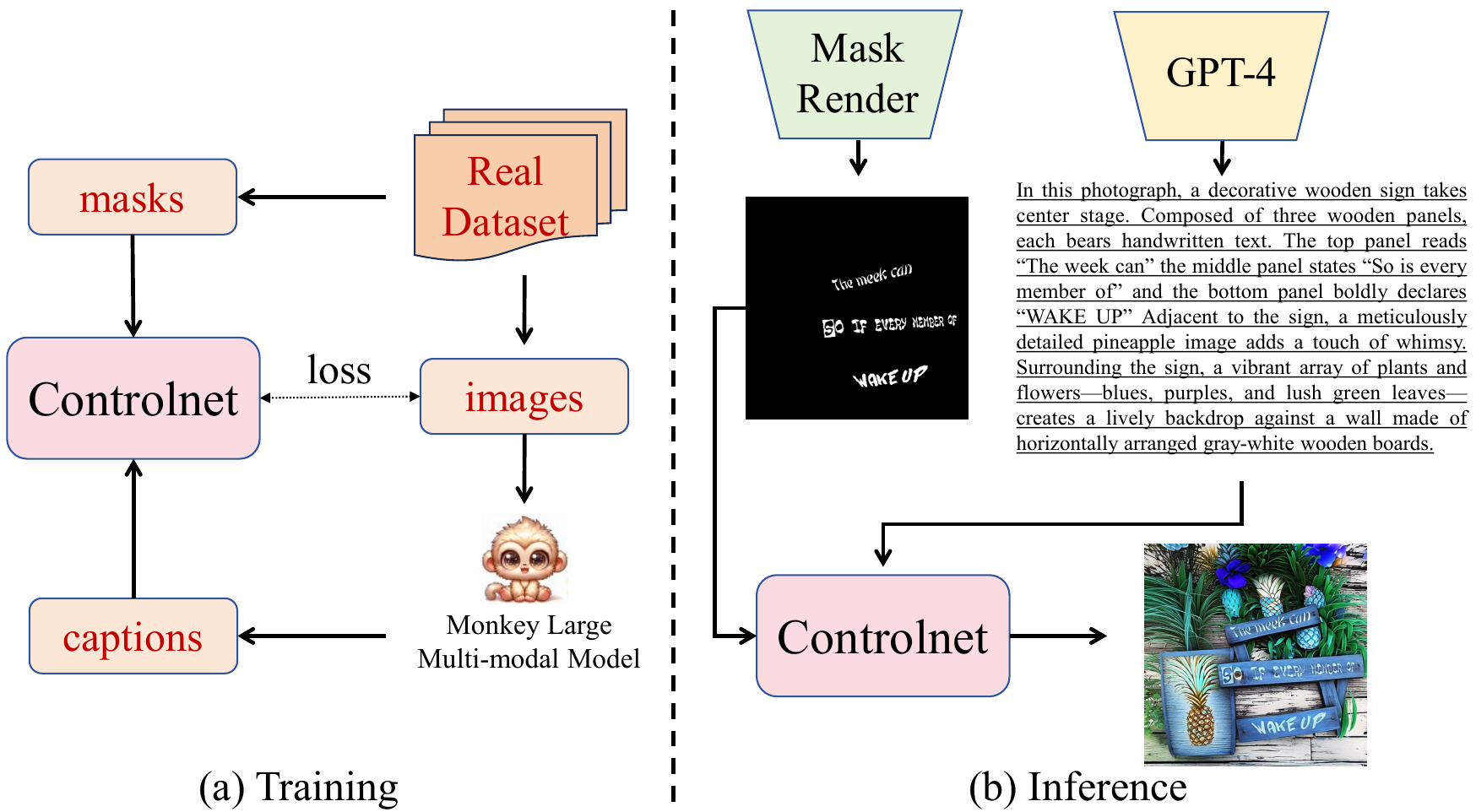}
\caption{(a) Training pipeline of ControlNet. (b) WAS-S data generation pipeline. }
\label{fig:pipeline}
\end{figure}

\subsection{WAS-S Synthetic Dataset Construction}
Fig.~\ref{fig:pipeline} shows the pipeline of generating synthetic text images. The core idea is that we build a text image generation model which can generate aligned text images from text masks and input prompts. To this end, we construct the training pipeline as illustrated in Fig.~\ref{fig:pipeline} (a). Specifically, we first generate diverse and informative captions from the text images in the training set of WAS-R to obtain training triplets <caption, mask, image>. Following that, we train a ControlNet~\cite{DBLP:conf/iccv/ZhangRA23} with these triplets for generating diverse images that are pixel-wisely aligned with the input text mask. During inference, as shown in Fig.~\ref{fig:pipeline} (b), we first construct diverse text masks using Mask Render, then use GPT-4 to extend the texts in the mask into a scene description caption. The constructed text mask and caption are send to our trained ControlNet to generate the synthetic text image. We describe the modeling details below.


\noindent\textbf{Training Pipeline.} During training, we construct the  dataset for training ControlNet~\cite{DBLP:conf/iccv/ZhangRA23} based on the training set of WAS-R. To this end, we obtain the image captions of existing training samples in WAS-R using the advanced multi-modal large language model called Monkey~\cite{li2024monkey}. Formally, let $I_t$ represent the real image in the training set of WAS-R, and $C_t$ denote the prompt generated from $I_t$, we have $C_t = Monkey(I_t)$.

Having obtained the <caption, text mask, text image> training triplets, we train a ControlNet that maps the input prompt and text mask to a text image. Our goal is that the outline of the artistic text in the generated image should be well aligned with the input text mask. Moreover, the contents and styles of the generated images should be diverse enough as guided by the input prompts. 


\noindent\textbf{Inference Pipeline.}  During inference, as illustrated in Fig.~\ref{fig:pipeline}(b), we first generate synthetic text masks denoted as $M_{syn}$ using our proposed \textbf{Mask Render} technique. Specifically, for each mask, we randomly select 1-7 phrases from the 20 newsgroups dataset~\cite{DBLP:conf/webi/AlbishreAL15} based on the word distribution of each image in the real dataset WAS-R. These phrases are consist of 1-5 consecutive words. Additionally, we apply a random rotation in the range of $-30^\circ \leq \phi \leq 30^\circ$ to each phrase. The size of each phrase is limited to match the general width of the entire image, and we position them randomly within the image boundaries. Besides, we use 250 artistic fonts.
Finally, an affine transformation is applied to each phrase to introduce skewness and distortion. 

\begin{figure}
\centering
\includegraphics[width=1\textwidth]{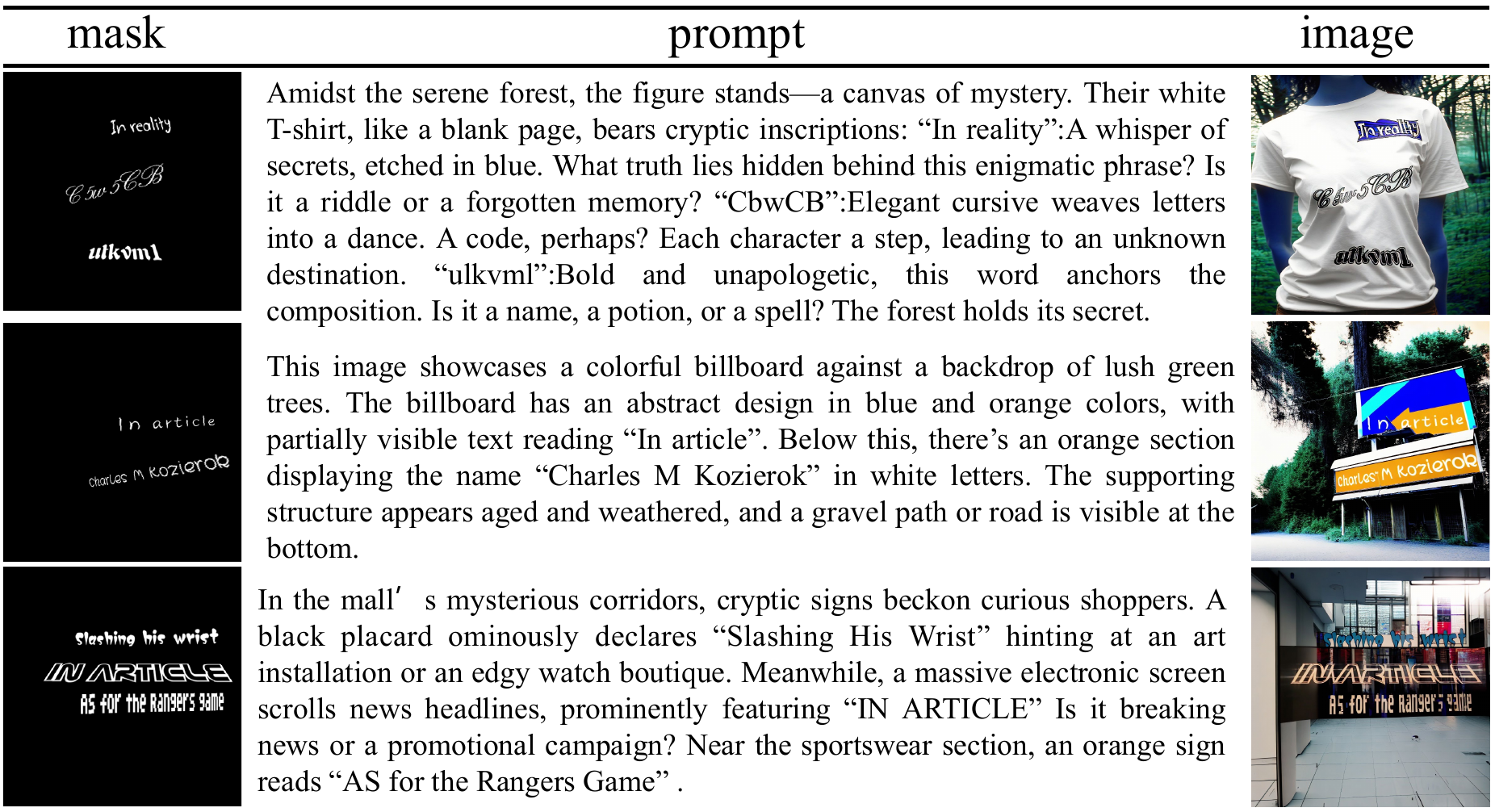}
\caption{The generated <mask, prompt, image> triplet. The left column is the generated masks. The middle column shows the prompt generated by GPT-4, imitating styles of the prompt in the training set. The right column is the final generated images.}
\label{fig:result}
\end{figure}

We use GPT-4~\cite{bubeck2023sparks} to generate the prompts corresponding to the synthetic text masks. Specifically, we ask GPT-4 to mimic the style of captions we generate from the training set in WAS-R, and synthesize new prompts. Next, we incorperate the text information in the synthetic mask into the generated prompt to obtain the final prompt. Formally, we have: $C_{syn} = GPT4(C_t, C_{M})$, where $C_{syn}$ is the caption we generate and $C_{M}$ is the text in the synthetic mask. Fig.~\ref{fig:result} shows examples of our synthetic prompts. 

Following the construction of synthetic text mask and synthetic prompts, we use the trained ControlNet to generate the final text image. We have $I_{syn} = ControlNet_{\theta}(C_{syn}, M_{syn})$, where $I_{syn}$ is the image we generate, $\theta$ denotes the trainable parameters in ControlNet. Fig.~\ref{fig:result} shows examples of the final synthesized <text mask, prompt, image> triplet. More details can refer to the supplementary materials.

\section{Methodology}
In this section, we introduce our artistic text segmentation model WASNet. We first present the overall architecture, followed by detailed descriptions of the local and global designs.

\begin{figure}
  \centering
   \includegraphics[width=\linewidth]{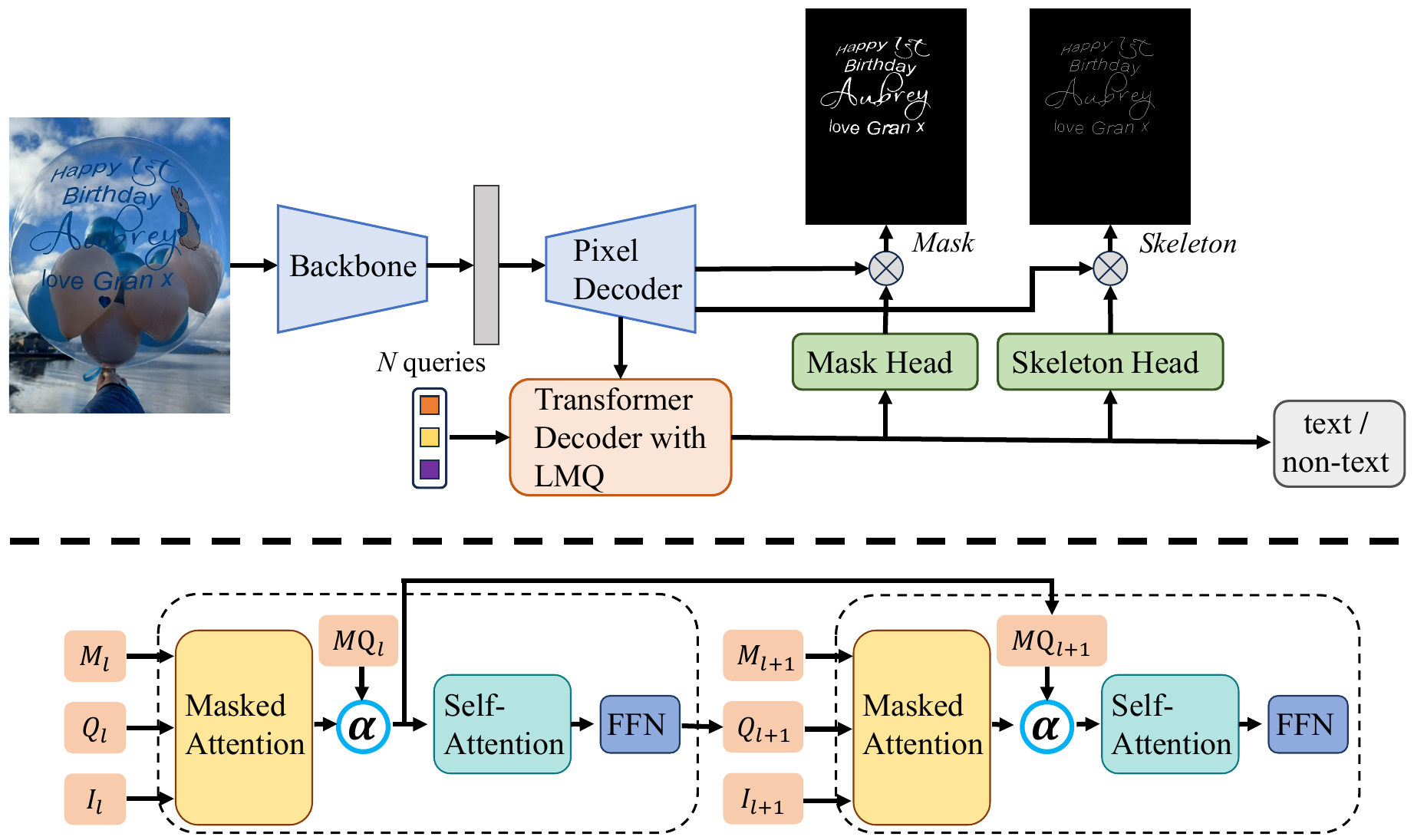}
   \caption{
   \textbf{Up:} The overall architecture of our WASNet.  \textbf{Down:} The Transformer decoder with layer-wise momentum query (LMQ).
   }
   \label{fig:Method}
\end{figure}

\subsection{Overall Architecture}
The overall framework of WASNet is shown in~\cref{fig:Method}. We take an excellent semantic segmentation model Mask2Former~\cite{cheng2022masked} as the meta-architecture. It is a mask classification architecture that directly predicts multiple binary masks and corresponding category labels, instead of performing per-pixel classification.
We add a skeleton-assisted head and improve the Transformer decoder with a mechanism of layer-wise momentum query. The backbone extracts low-resolution features from an image. The pixel decoder upsamples the image features and generates a feature
pyramid. The multi-scale features are fed into the Transformer decoder, with each resolution corresponding to each layer's input of the decoder. Besides, each layer of the Transformer decoder also receives the mask prediction and query generated from the previous layer as input. Finally, the mask head and the skeleton head generate binary mask and skeleton predictions respectively, by decoding the per-pixel embeddings from the pixel decoder and object queries from the Transformer decoder. The ground truth of the skeleton is obtained by thinning the binary mask labels through a skeleton extraction algorithm~\cite{zhang1984fast}.

\subsection{Transformer Decoder with Layer-wise Momentum Query}

Artistic text segmentation faces the challenge of the local stroke shapes being flexible and changeable. Due to designers using hundreds of different artistic fonts and applying various text effects, the local strokes of the same character can differ significantly. This results in some slender strokes spanning across other areas, as well as twisted ligatures leading to complex text edges.  In contrast, normal scene text typically utilizes regular printed fonts without special designs, and the stroke shapes are almost invariant. Therefore, it is necessary for the decoder to pay attention to these special local strokes.

First, we use the masked attention mechanism~\cite{cheng2022masked}, constraining cross-attention
to within the local text mask region for each query, instead of attending to the full feature map. This mechanism can be expressed as:
\begin{equation}
\mathbf{MA}_l=\operatorname{softmax}\left(\mathbf{M}_l + \mathbf{Q}_l\mathbf{I}_l^{\mathrm{T}}\right) \mathbf{I}_l,
\end{equation}
where $l$ is the layer index, and $\mathbf{Q}_l$ is the input queries. $\mathbf{I}_l \in \mathbb{R}^{H_l W_l \times C}$ is the image feature input to the $l$-th layer, which comes from the feature pyramid of the pixel decoder. $H_l$ and $W_l$ indicate the spatial resolution of the image feature and $C$ is the feature dimension. $\mathbf{M}_l$ is transformed from the binary mask output of the previous layer, with the value of the text region being $0$ and the value of the non-text region being $-\infty$~\cite{cheng2022masked}.
$\mathbf{MA}_l$ is the output of the masked attention module. We have omitted the residual connection and normalization here. 

Furthermore, since the masks predicted by each layer are different, the previous layers yield coarse masks that may include special-shaped stroke regions. However, the subsequent layers are inclined to predict more precise regions of regular strokes, overlooking those local special regions. Therefore, in order to prevent the model's attention from being quickly confined to regular regions, we design a mechanism of Layer-wise Momentum Query (LMQ). The momentum superposition of the masked queries from the current and previous layers is input to the self-attention module before the module gathers contextual information. ~\cref{equ:LMQ} illustrates this mechanism.
\begin{equation}
\label{equ:LMQ}
\mathbf{MQ}_{l+1}=\alpha\mathbf{MQ}_{l}+(1-\alpha)\mathbf{MA}_{l},
\end{equation}
where $\alpha \in[0,1)$ is a momentum coefficient. $\mathbf{MQ}$ is the momentum query that is input to the self-attention module. We ultimately use this decoder in WASNet with layer-wise momentum query.

\subsection{Skeleton-Assisted Head}
Different from regular text and general objects, the global topological structure of artistic text is very complex, and there are many holes and intricate connections inside. This presents new challenges for the segmentation task. The model needs to capture the global structure of the text object rather than just a region. Inspired by DeepSkeleton~\cite{shen2017deepskeleton} and DeepFlux~\cite{wang2019deepflux}, 
we found that the skeleton is an effective representation to describe the shape and topology of text because it can extract the central axis of the object. Therefore, we use skeletons to assist text segmentation.

As shown in~\cref{fig:Method}, we add a skeleton-assisted head to WASNet, enabling the model to simultaneously predict the mask and the skeleton, thus endowing it with the capability to perceive the global topological structure. Since the binary mask is a finely annotated label for semantic segmentation, the ground truth for the skeleton can be obtained by processing the mask with the classic Zhang-Suen~\cite{zhang1984fast} skeleton extraction algorithm. The algorithm progressively removes pixels that satisfy certain template structural conditions through an iterative process, until no more pixels meeting the conditions are deleted.

We use the binary cross-entropy loss and the dice loss~\cite{milletari2016v} for our skeleton loss and mask loss:
\begin{equation}	\mathcal{L}_{skeleton}=\mathcal{L}_{mask}= \lambda_{ce}\mathcal{L}_{\mathrm{ce}} + \lambda_{dice} \mathcal{L}_{\mathrm{dice}}.
\end{equation}
We set $\lambda_{ce}=\lambda_{dice}=5$. The final loss is the combination of skeleton loss, mask loss, and classification loss:
\begin{equation}	\mathcal{L}_{final}=\mathcal{L}_{skeleton}+\mathcal{L}_{mask}+ \lambda_{cls} \mathcal{L}_{\mathrm{cls}},
\end{equation}
where $\lambda_{cls}=2$ for
predictions matched with labels and $0.1$ for predictions that have not been matched with any labels.

During the inference phase, it is unnecessary to output the predictions of the skeleton. Therefore, we follow the post-processing method in~\cite{cheng2021per} to obtain the final output of text semantic segmentation.

\setlength{\tabcolsep}{4pt}
\begin{table}
\begin{center}
\caption{Performance comparison with other methods on WAS dataset. * TextFormer trains a text detection module using additional bounding box labels. “pre-train” indicates that the model was firstly trained on WAS-S and then fine-tuned on WAS-R.}
\label{table:was}
\begin{tabular}{l|c|cc}
\hline\noalign{\smallskip}
\multirow{2}*{Methods} & \multirow{2}*{Venue} & \multicolumn{2}{c}{WAS} \\
& & fgIoU & F-score \\
\noalign{\smallskip}
\hline
\noalign{\smallskip}
PSPNet~\cite{pspnet} & CVPR'17 & 71.15 & 0.831 \\
DeepLabV3+~\cite{deeplabv3+} & CVPR'18 & 79.65 & 0.887 \\
OCRNet~\cite{ocr_seg} & ECCV'20 & 79.06 & 0.883 \\
SegFormer~\cite{xie2021segformer} & NeurIPS'21 & 79.46 & 0.886\\
TexRNet~\cite{xu2021rethinking} & CVPR'21 & 77.19 & 0.850\\
DDP~\cite{ji2023ddp} & ICCV'23 & 81.07 & 0.896\\
TextFormer*~\cite{yu2023scene} & ACM MM'23 & 80.12 & 0.889\\
\noalign{\smallskip}
\hline
\noalign{\smallskip}
Mask2Former~\cite{cheng2021per} & NeurIPS'21 & 80.21 & 0.890\\
WASNet (ours) & - & 82.11 & 0.901 \\
Mask2Former (pre-train) & - & 82.42 & 0.902\\
WASNet (pre-train) & - & \textbf{84.18} & \textbf{0.913} \\
\hline
\end{tabular}
\end{center}
\end{table}

\section{Experiments}

\subsection{Implementation Details}
Our experiments are mainly based on the MMSegmentation~\cite{mmseg2020} toolbox. The overall hyperparameter configuration is the same as~\cite{cheng2022masked}. The pixel decoder is a multi-scale deformable attention Transformer~\cite{DBLP:conf/iclr/ZhuSLLWD21} with 6 layers. The Transformer decoder consists of 9 layers, each with an auxiliary loss.
We use the AdamW~\cite{DBLP:conf/iclr/LoshchilovH19} optimizer and the poly~\cite{DBLP:journals/pami/ChenPKMY18} learning rate schedule with an initial learning rate of $10^{-4}$ and a weight decay of 0.05. The data augmentation strategies include random scale jittering, random color jittering, random cropping as well as random horizontal flipping. We use a crop size of $512\times512$ and a batch size of 16. The models are trained with 8 RTX4090 GPUs. 
If the model is only trained on the real dataset, we set the number of iterations to 100k. If the model needs to be pre-trained on the synthetic dataset WAS-S, we first pre-train the model for 50k iterations and then fine-tune it on the real dataset for 50k iterations.
For the momentum coefficient $\alpha$ in~\cref{equ:LMQ}, we set $\alpha=0.8$ by default.
Following the previous text segmentation methods~\cite{yu2023scene,xu2021rethinking}, we use foreground (text) Intersection-over-Union (fgIoU) as the major metric and F-score measurement on foreground pixels as the auxiliary metric. 

\subsection{Results of Artistic Text Segmentation}
\label{sec:ats}

To verify the superiority of our method in the task of artistic text segmentation, we trained several representative models on our WAS-R dataset, including six semantic segmentation models and two text segmentation models. We use the officially released code for TexRNet~\cite{xu2021rethinking}, DDP~\cite{ji2023ddp}, and TextFormer~\cite{yu2023scene}, and the code reproduced by MMSegmentation~\cite{mmseg2020} for other models. For a fair comparison, we did not apply the character-level glyph discriminator for TexRNet.

The experimental results in~\cref{table:was} indicate that our WASNet outperforms all of these advanced models. Moreover, when we train the baseline models and WASNet with the synthetic dataset WAS-S, their performance can be further improved. Our final results have achieved a significant SOTA performance.

\subsection{Results of Scene Text Segmentation}
To further verify the generalizability of WASNet, we also conducted experiments on three publicly available scene text segmentation datasets~\cite{cocots,totaltext,xu2021rethinking}, as shown in~\cref{tab:sts_benchmark}. We can draw the same conclusion as in~\cref{sec:ats} regarding the effectiveness of WASNet and our synthetic dataset. It is worth mentioning that character-level annotations were used to train TexRNet on TextSeg. Extra bounding box labels were used to train the text detection module of TextFormer on all three datasets. However, we only use binary mask labels of the full images. Despite this, we still achieved competitive or state-of-the-art results. Due to the highly inaccurate annotation quality of COCO\_TS~\cite{cocots} and the fact that Total-Text~\cite{totaltext} contains only 300 test images, the conclusions drawn from the evaluation results of the models on these two datasets may be inconsistent.

\setlength{\tabcolsep}{4pt}
\begin{table}[htbp]
\centering
\caption{Performance comparison with other methods on three publicly available scene text segmentation datasets. * TextFormer trains a text detection module using additional bounding box labels. “pre-train” indicates that the model was firstly trained on WAS-S and subsequently fine-tuned on the specific datasets.}
\label{tab:sts_benchmark}
\begin{tabular}{l|cc|cc|cc}
\hline
\multirow{2}*{Methods}  & \multicolumn{2}{c|}{COCO\_TS~\cite{cocots}} & \multicolumn{2}{c|}{Total-Text~\cite{totaltext}} & \multicolumn{2}{c}{TextSeg~\cite{xu2021rethinking}} \\
& fgIoU & F-score & fgIoU & F-score & fgIoU & F-score \\
\hline
PSPNet~\cite{pspnet}  & - & - & - & 0.740 & - & - \\
SMANet~\cite{cocots}  & - & - & - & 0.770 & - & - \\
DeepLabV3+~\cite{deeplabv3+}  & 72.07 & 0.641 & 74.44 & 0.824 & 84.07 & 0.914 \\
HRNetV2-W48~\cite{hrnet}  & 72.07 & 0.641 & 74.44 & 0.824 & 85.03 & 0.914 \\
OCRNet~\cite{ocr_seg}  & 69.54 & 0.627 & 76.23 & 0.832 & 85.98 & 0.918 \\
SegFormer~\cite{xie2021segformer}  & 63.17 & 0.774 & 73.31 & 0.846 & 84.59 & 0.916 \\
TexRNet~\cite{xu2021rethinking}  & 72.39 & 0.720 & 78.47 & 0.848 & 86.84 & 0.924 \\
DDP~\cite{ji2023ddp}  & 70.04 & 0.824 & 72.55 & 0.841 & 84.37 & 0.915 \\
TextFormer*~\cite{yu2023scene}  & \textbf{73.40} & \underline{0.847} & \textbf{82.10} & \textbf{0.902} & \underline{87.11} & \underline{0.931} \\ 
\hline
Mask2Former~\cite{cheng2021per} (baseline)  & 70.03 & 0.823 & 75.54 & 0.832 & 84.95 & 0.911 \\
WASNet (ours)  & 71.10 & 0.830 & 77.26 & 0.840 & 86.56 & 0.921 \\ 
Mask2Former (pre-train)  & 70.89 & 0.830 & 78.05 & 0.851 & 86.00 & 0.919 \\
WASNet (pre-train)  & \underline{73.28} & \textbf{0.848} & \underline{79.30} & \underline{0.863} & \textbf{87.42} & \textbf{0.932} \\
\hline
WASNet (WAS dataset)  & 69.22 & 0.817 & 75.39 & 0.836 & 84.26 & 0.906 \\
\hline
\end{tabular}
\end{table}

Furthermore, we directly evaluate the performance of WASNet on the three datasets using the model trained on the synthetic and real WAS datasets, as shown in the last row of~\cref{tab:sts_benchmark}. Note that the results in this row have not been fine-tuned on specific datasets but are still competitive. Therefore, to simplify the experimental paradigm and evaluation process of text segmentation models, we encourage researchers to train on WAS and test directly on other datasets.

\begin{minipage}{\textwidth}

\begin{minipage}[t]{0.45\textwidth}
\makeatletter\def\@captype{table}
\begin{center}
\caption{Ablation study on our proposed modules and datasets.}\label{tab:abl}
\begin{tabular}{l|cc}
\hline\noalign{\smallskip}
\multirow{2}*{Methods} & \multicolumn{2}{c}{WAS} \\
& fgIoU & F-score \\
\noalign{\smallskip}
\hline
\noalign{\smallskip}
Baseline~\cite{cheng2021per} & 80.21 & 0.890\\
\noalign{\smallskip}
\hline
\noalign{\smallskip}
+ LMQ & 80.95 & 0.898 \\
+ Skeleton  & 82.11 & 0.901\\
+ WAS-S & \textbf{84.18} & \textbf{0.913} \\
\hline
\end{tabular}
\end{center}
\end{minipage}
\begin{minipage}[t]{0.48\textwidth}
\makeatletter\def\@captype{table}
\begin{center}
\caption{Ablation study on datasets.}\label{tab:data}
\begin{tabular}{l|cc}
\hline\noalign{\smallskip}
\multirow{2}*{Dataset} & \multicolumn{2}{c}{WAS} \\
& fgIoU & F-score \\
\noalign{\smallskip}
\hline
\noalign{\smallskip}
WAS-S & \textbf{84.18} & \textbf{0.913} \\
\noalign{\smallskip}
\hline
\noalign{\smallskip}
5w images& 83.25 & 0.906 \\
20w images& 84.01 & 0.912\\
BLIP2~\cite{DBLP:conf/icml/0008LSH23}  & 83.07 & 0.906\\
1000 fonts & 82.35 & 0.902 \\
\hline
\end{tabular}
\label{sample-table}
\end{center}
\end{minipage}
\end{minipage}

\setlength{\tabcolsep}{4pt}
\begin{table}
\begin{center}
\caption{Ablation study on the momentum coefficient.} \label{mo}
\begin{tabular}{c|ccccccc}
\hline\noalign{\smallskip}
$\alpha$ & 0.9 & 0.8 & 0.6 & 0.5 & 0.4 & 0.2 & 0.1 \\
\noalign{\smallskip}
\hline
\noalign{\smallskip}
fgIoU & 82.03 & \textbf{82.11} & 82.06 & 81.87 & 81.92 & 81.36 & 81.31 \\
\hline
\end{tabular}
\end{center}
\end{table}

\subsection{Ablation Study}
In this section, we conduct the ablation study on the artistic text segmentation dataset WAS.
We first validate the effectiveness of our proposed modules and the synthetic dataset. As shown in~\cref{tab:abl}, as we gradually apply the design of LMQ and the skeleton to the baseline, the performance of WASNet is incrementally improved. Pre-training WASNet on WAS-S can continue to enhance the performance of artistic text segmentation.

Therefore, the synthetic dataset is an important contribution of this paper. We conduct ablation experiments regarding some synthetic details of the dataset in~\cref{tab:data}. It is crucial to control the number of synthesized mask-image pairs. Less data will weaken performance, but more data will lead to a performance plateau.
The reasons will be analyzed in~\cref{lim}.
We also use other large multi-modal model BLIP2~\cite{cheng2021per} to generate the image captions, but the performance is limited. This is because the overall performance of BLIP2 is inferior to Monkey~\cite{li2024monkey} we used. Besides, we applied more fonts to generate masks, but the performance actually decreased. The dataset of 1000 fonts includes a large number of regular fonts, which reduces the learning difficulty of the dataset.

Furthermore, we explored the impact of different momentum coefficient values $\alpha$ on the performance of WASNet in~\cref{mo} and found that the approximate optimal value is 0.8. A coefficient that is too large can cause the model to be overly influenced by the coarse predictions from earlier layers. A coefficient that is too small diminishes the positive effect of momentum queries.


\subsection{Further Analysis}
\begin{figure}
  \centering
   \includegraphics[width=0.86\linewidth]{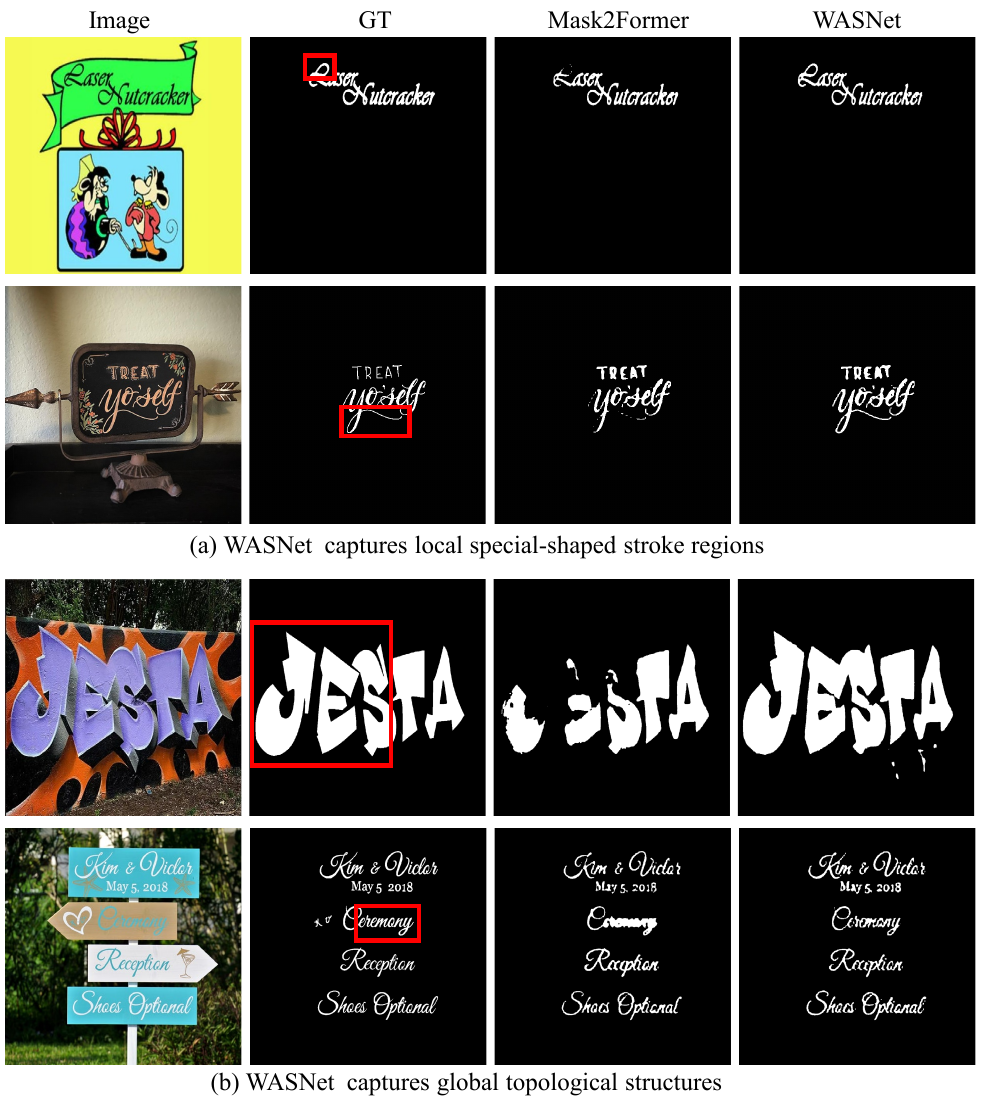}
   \caption{
   Qualitative comparison between the baseline model Mask2Former~\cite{cheng2021per} and our WASNet. The two innovations of our method alleviate the two main problems of artistic text segmentation respectively.
   }
   \label{fig:vis}
\end{figure}

To further verify the effectiveness of our proposed WASNet, we visualize the inference outputs of our baseline model Mask2Former~\cite{cheng2021per} and WASNet in~\cref{fig:vis}. According to~\cref{fig:vis} (a), it is evident that WASNet can capture special-shaped stroke regions such as slender tails or twisted ligatures. This is attributed to our  Transformer decoder with the layer-wise momentum query. Additionally, according to~\cref{fig:vis} (b), WASNet exhibits good scale adaptability. It can achieve fine results for both large-scale and small-scale text with complex structures. This is because the skeleton-assisted head can obtain the global topological structure of the text through the thinning operation, guiding fine segmentation. 

Once accurate text stroke masks are obtained, downstream text-related generative tasks can demonstrate excellent results. The application effects of text removal, text background replacement, and text style transfer are shown in the supplementary materials.

\subsection{Limitation}
\label{lim}
Although the proposed synthetic dataset can improve the performance of text segmentation models, the enhancement is limited and does not \emph{significantly} increase. Even when we further increased the amount of synthetic data, the performance remained unchanged. This could be caused by the bottleneck encountered in the diversity and realism of the synthetic images. In the future, we are considering designing more advanced generative models.

\section{Conclusion}
The paper focuses on a new challenging task of artistic text segmentation. We propose a real dataset for this task to train the models and benchmark the performance. We also construct a synthetic dataset to further enhance the accuracy and generalization ability. In order to meet the challenges of this task, we introduce the layer-wise momentum query to handle the changeable local strokes and the skeleton-assisted head to capture the complex global structure. Experimental results have demonstrated the effectiveness and superiority of our method in the tasks of artistic text segmentation and scene text segmentation. We hope that more researchers can focus on this task in the future and that the dataset we propose can change the experimental paradigm of text segmentation.

\section*{Acknowledgements}
This work was supported by the National Science Fund for Distinguished Young Scholars of China (Grant No.62225603).

%
%
\bibliographystyle{splncs04}
\bibliography{main}

\newpage
\appendix
\section{More Details on Synthetic Dataset Construction}
As stated in the paper, we first construct the training pipeline for a text image generation model, learning to generate text images spatially aligned with text masks. Then we construct an inference pipeline to input new masks and prompts into the trained generation model, generating new text images. Here we add the details of prompt generation in the training and inference pipelines.
\subsection{Training Pipeline}
\begin{figure}
  \centering
   \includegraphics[width=\linewidth]{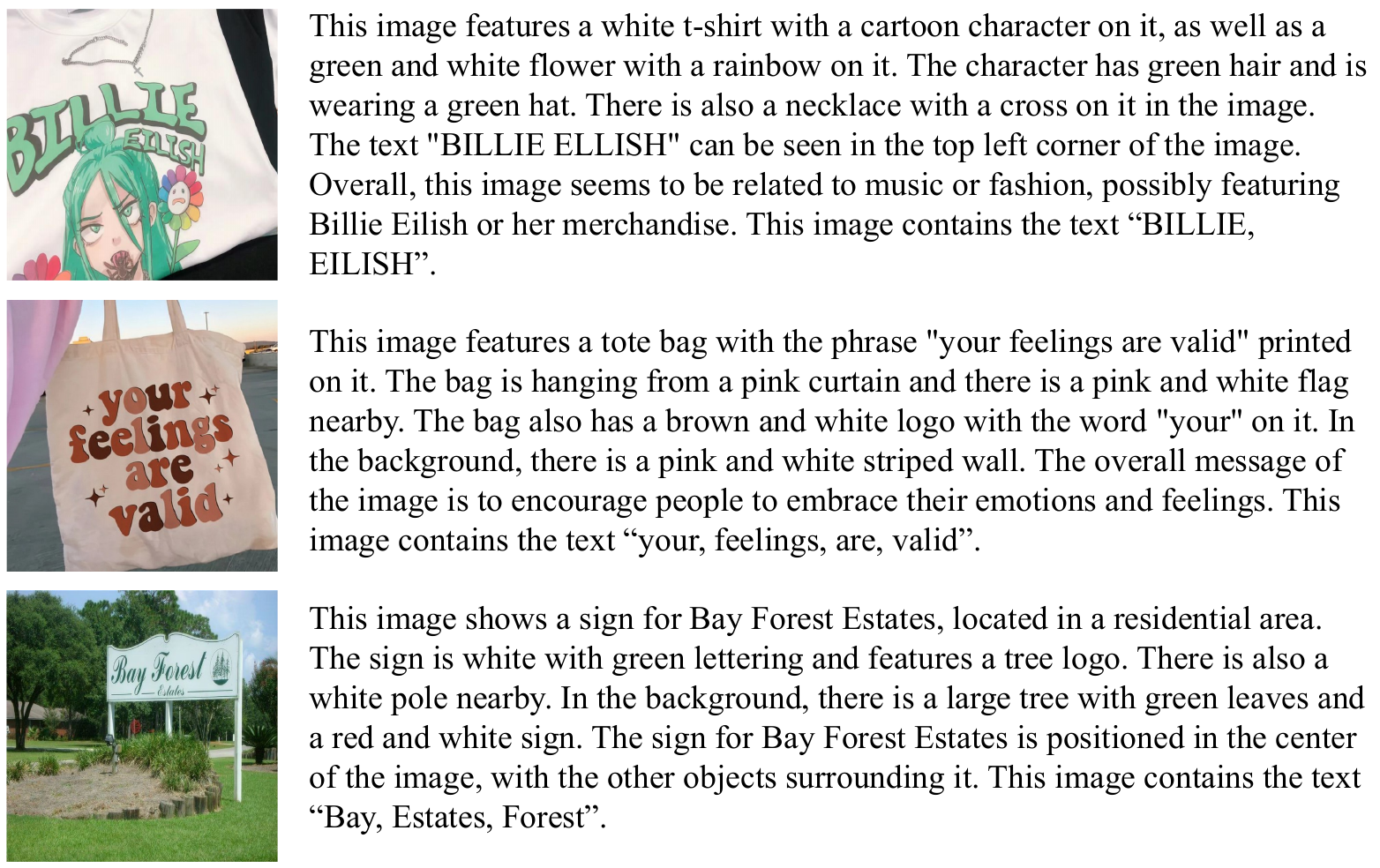}
   \caption{
   The captions generated by Monkey~\cite{li2024monkey} for training.
   }
   \label{fig:supp_monkey}
\end{figure}

To train the text image generation model such as ControlNet~\cite{DBLP:conf/iccv/ZhangRA23}, it is necessary to obtain training data of <caption, text mask, text image> triplets. Text masks and text images are from our proposed real dataset. Captions should be detailed descriptions of the text images. To this end, we utilize a large multi-modal model, Monkey~\cite{li2024monkey}, to caption the images. Monkey is an open-source model and can handle vision-language tasks with high-resolution input and detailed scene understanding. It performs well on Image Captioning and various Visual Question Answering (VQA) tasks. Therefore, we feed a text image and a prompt \emph{``generate the detailed caption in English''} to Monkey and let it output a detailed description. The examples of the generated captions are shown in~\cref{fig:supp_monkey}. We found that, in many cases, Monkey is able to recognize and describe the text in images. To ensure the accuracy of the descriptions and to highlight the importance of the text, we add a sentence after each caption: \emph{This image contains the text ``text in the image''.}

\begin{figure}
  \centering
   \includegraphics[width=\linewidth]{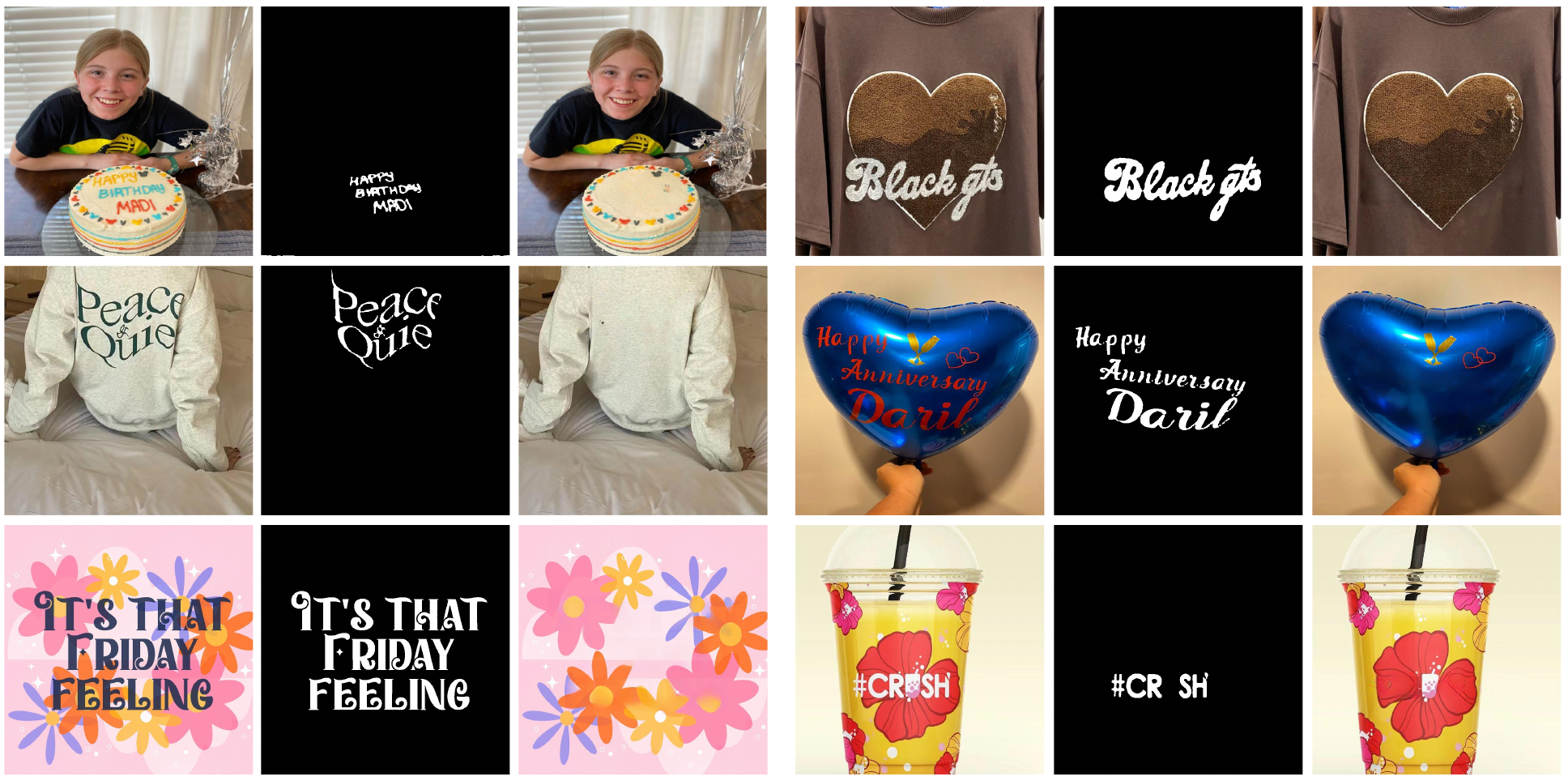}
   \caption{
   Text removal visualization using predicted text masks from our WASNet and inpainting model LaMa~\cite{Suvorov_2022_WACV}. Each sample includes the original image, the predicted mask, and the text removal result from left to right.
   }
   \label{fig:supp_removal}
\end{figure}

\subsection{Inference Pipeline}
During the inference phase, we first need to produce new binary masks of text through the Mask Render introduced in the paper. Moreover, it is crucial to generate new prompts that describe more complex scenes. Combining the masks with rich descriptions of scenes, the trained model can generate new and realistic text images. We use GPT-4~\cite{bubeck2023sparks} to generate the prompts. To ensure that the new prompts and the training prompts are in the same domain, and avoid domain gaps in the images generated by the model, we first provide GPT-4 with 50 caption examples produced by Monkey. Then we ask GPT-4 to mimic the style of these captions and synthesize new prompts. The instruction is \emph{Please follow the above caption examples and generate a similar caption, which must contain some double-quoted spaces `` ''.} Next, we insert the text corresponding to each new mask into the quotation marks in the new prompt, forming the final prompt. The generated <prompt, mask, image> triplet is shown in~\cref{fig:result}.

\section{Applications}
\subsection{Text Removal}
Text removal refers to the process of erasing or deleting text regions from an image. The finer the text mask, the better the erasing performance, as it preserves more background pixels. Therefore, stroke-level text segmentation can greatly benefit this task. Text removal is essentially an image inpainting task, so LaMa~\cite{Suvorov_2022_WACV} is employed and the results are shown in~\cref{fig:supp_removal}. 

\begin{figure}
  \centering
   \includegraphics[width=\linewidth]{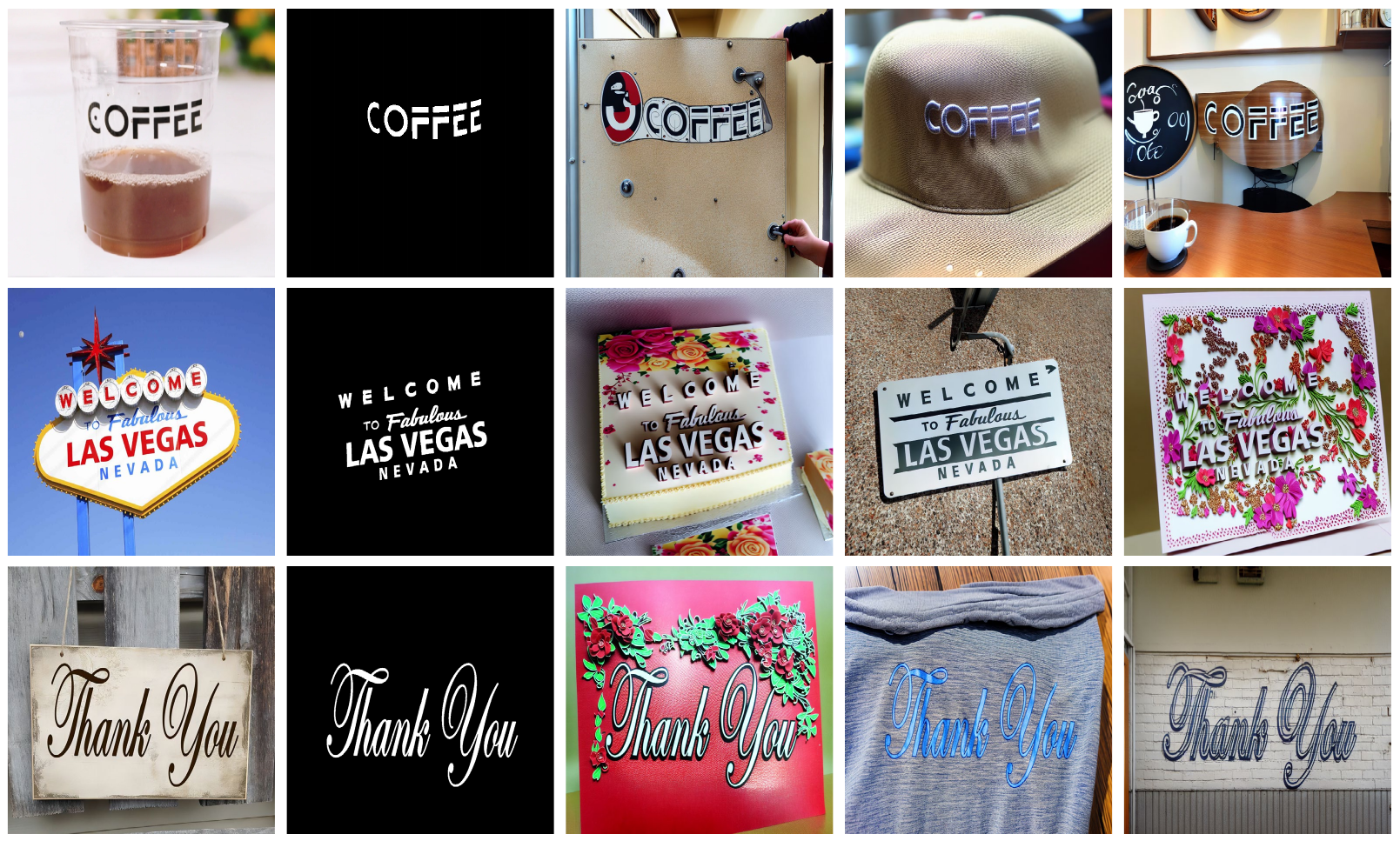}
   \caption{
   Visualization for text background replacement. The first column displays the original images. The second column shows the predicted masks from our WASNet. The remaining three columns show the images whose backgrounds have been replaced.
   }
   \label{fig:supp_replace}
\end{figure}

\subsection{Text Background Replacement}
Once we have obtained the fine mask of the text, we can freely replace the background of the image, embedding the text into various scenes. We use ControlNet~\cite{DBLP:conf/iccv/ZhangRA23} to replace the background and~\cref{fig:supp_replace} presents the results.

\begin{figure}
  \centering
   \includegraphics[width=\linewidth]{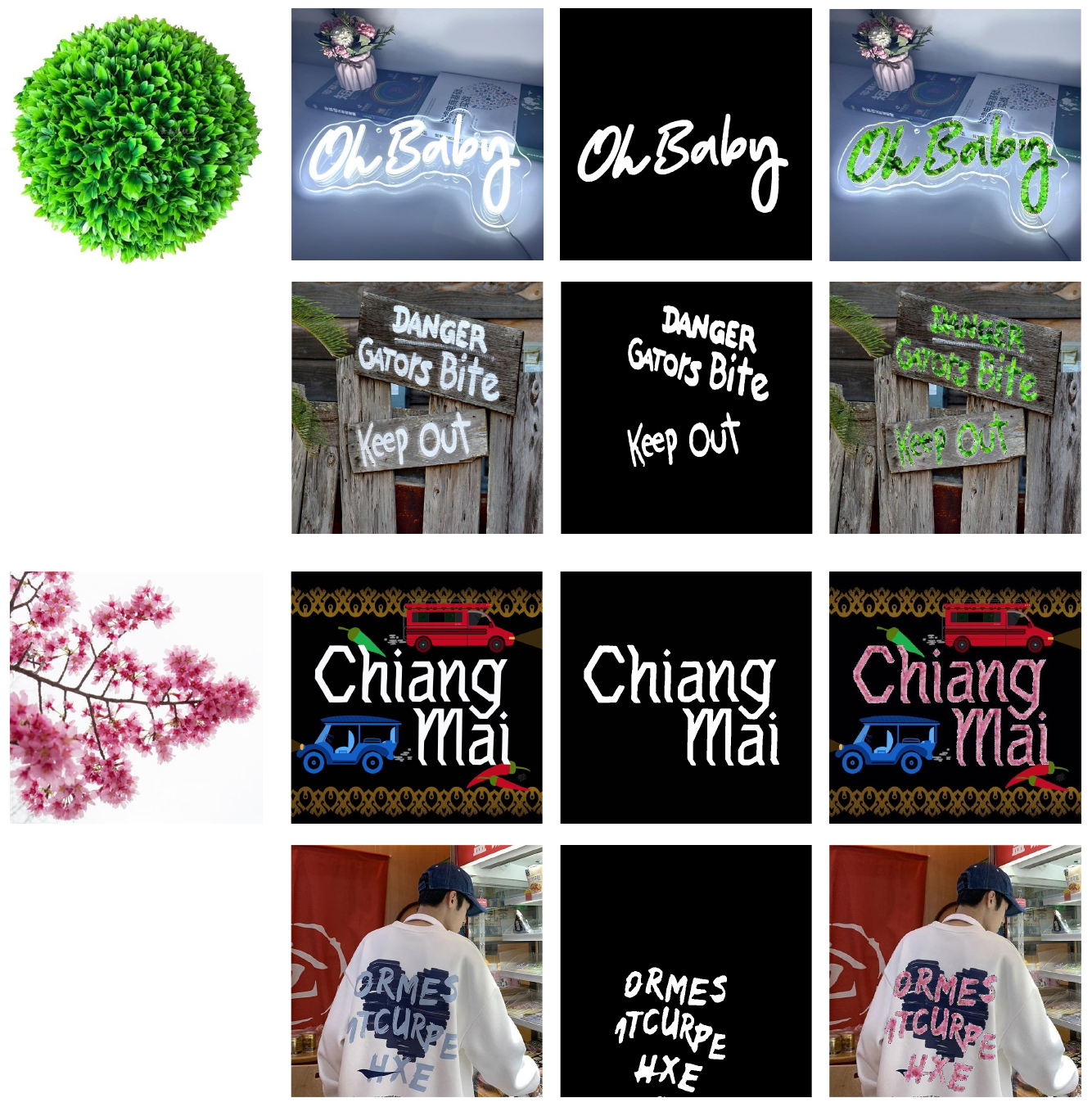}
   \caption{
   Visualization for text style transfer. 
   The first column displays two style reference images. The second and third columns show the original images and the predicted text masks. The last column displays the images with stylized text.
   }
   \label{fig:supp_style}
\end{figure}

\subsection{Text Style Transfer}
Text style transfer is a task that renders text in natural images into artistic text according to a style reference image while keeping the text content unchanged. It usually relies on accurate text masks. We use Intelligent Typography~\cite{mao2022intelligent} as the style transfer model and input the predicted text masks to it. The stylized text is shown in the last column of~\cref{fig:supp_style}.
\end{document}